\documentclass[11pt,twocolumn]{scrartcl}

\usepackage{casa_conf}	
\usepackage{graphicx}	
\usepackage{bm}
\usepackage{amssymb}
\usepackage[percent]{overpic}
\usepackage{amsmath}
\usepackage{gensymb}
\usepackage{tabularx}
\usepackage{float}

\title{Robust Loss Functions for Object Grasping under Limited Ground Truth}

 \author{Yangfan Deng\\
 		School of Mathematical Sciences\\
        Ocean University of China\\
        dengyangfan@stu.ouc.edu.cn\\
        \and
        Mengyao Zhang\\
        School of Mathematical Sciences\\
        Ocean University of China\\
        MengYaoZhangOUC@163.com
        \and
        Yong Zhao*\\
        School of Mathematical Sciences\\
        Ocean University of China\\
        zhaoyong@ouc.edu.cn}

\begin{document}

\maketitle

\begin{abstract}	
Object grasping is a crucial technology enabling robots to perceive and interact with the environment sufficiently. However, in practical applications, researchers are faced with missing or noisy ground truth while training the convolutional neural network, which decreases the accuracy of the model. Therefore, different loss functions are proposed to deal with these problems to improve the accuracy of the neural network. For missing ground truth, a new predicted category probability method is defined for unlabeled samples, which works effectively in conjunction with the pseudo-labeling method. Furthermore, for noisy ground truth, a symmetric loss function is introduced to resist the corruption of label noises. The proposed loss functions are powerful, robust, and easy to use. Experimental results based on the typical grasping neural network show that our method can improve performance by 2 to 13 percent.
\end{abstract}
\linebreak
\linebreak


\section{Introduction}
Although manipulating objects is a simple task for humans, it is still a challenging problem for robots to achieve effective grasping of any object. It is widely accepted that object grasping is one of the basic operations to achieve robot control. Solving this problem will promote the use of robotics in industrial cases such as part assembly and binning.

In the last decade, convolutional neural network has achieved significant success on detection, classification and regression tasks. It is reasonable for researchers to introduce CNN into object grasping. Typical steps utilizing deep learning for grasp generation are shown in Figure 1. The input scenes are cluttered with multiple target objects occluding each other. Therefore, the researchers need to segment the target objects firstly and utilize 6D pose estimation techniques to accurately estimate the positions of the objects. Then the grasp generation network can generate grasp candidates for the target objects. Finally, grasps for the target objects need to be filtered by the evaluation network including collision detection, robustness testing and success rate filtering and etc.

\begin{figure*}[t] 
	\centering
	\includegraphics[width=\textwidth]{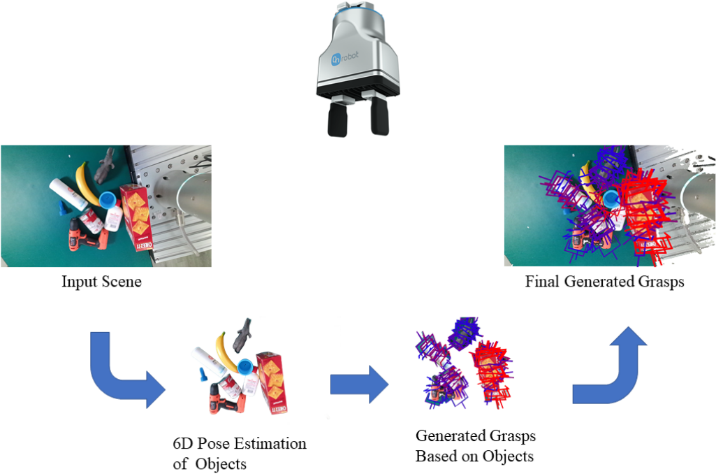} 
	\caption{The most classical steps for generating grasps. After receiving the scene data as input, the neural network will segment the objects firstly and the position of the target objects will be determined by 6D pose estimation technique. Subsequently, grasps based on the target objects can be generated. Finally, all generated grasps candidates need to be evaluated in order to find the best ones, and the evaluation methods includes collision detection, success rate of grasping and etc. The gripper of the robot is capable of performing the grasp following the generated grasps.} 
	\label{fig:my_label} 
\end{figure*}

However, object grasping problem still remains complicated to solve. The first challenge is the cluttering scenes existed in the environments where other similar objects may greatly decrease the successful possibility of grasping the target object. The second challenge is the generalization ability of the model. In the industrial environment, the object grasping algorithm should not only achieve accurate grasp of objects in the training dataset, but also generate effective grasp for the unseen objects. Furthermore, the most serious and important challenge is the validity and quality of data. Training data obtained in an industrial environment is probably facing with the problems of missing or noisy ground truth, which will reduce the performance of the neural network. Missing ground truth probably result in the risk of overfitting and lead to training instability. Simultaneously, due to the lack of the ground truth, the grasp evaluation network is unable to evaluate the performance of grasps accurately. As for noisy ground truth, it potentially misleads the training model into incorrect training, and correcting these erroneous labels requires a huge amount of time and resources.

Our proposed loss functions contribute to solve the problems of missing or noisy ground truth. Firstly, to deal with missing ground truth, a new predicted category probability method is introduced for unlabeled samples to generate better pseudo-labels. In particular, this new formulation conveys more information of missing labels. Secondly, this idea is extended to noisy ground truth, and further construct a symmetric loss function to reduce the corruption of label noises. A large number of experiments show that the performance of the algorithm can be greatly improved.
To the best of our knowledge, it is the first work to address missing or noisy ground truth in object grasping. We develop new loss functions, which are very effective and can be incorporated into existing grasping neural networks easily.

\section{Related work}
In this part, object grasping algorithms based on RGB or RGB-D data are reviewed firstly. Then several representative grasp algorithms related to point cloud are introduced. Subsequently, typical methods for missing ground truth and noisy ground truth are discussed.

\textbf{Grasping algorithms utilizing 2D and 2.5D data} Since Ian Lenz systematically introduced deep learning techniques into object grasp algorithms~\cite{1, 2}, more and more researchers were inspired to work on relevant research. There are primarily two directions: improving the accuracy of the grasp algorithm or the ability to handle with the cluttered environment. Researchers typically enhanced the accuracy by introducing advanced deep learning network into object grasp algorithms~\cite{3, 4, 5}. GR-ConvNet~\cite{3} utilized ResNet in the network, and FCGDN~\cite{4} introduced oriented anchor box to generate grasp candidates. GraspNet~\cite{5} based on the AutoEncoder could extract more accurate feature maps from the input images. As for handling with the clutter, many researchers interested in applying robotic grasping into industrial scenarios had made significant contributions to this field~\cite{6, 7, 8, 9, 10}. Sergey et al.~\cite{6} developed a grasp success rate prediction model and integrated it with Cross-Entropy Method algorithm to achieve the continuous control. Gualtieri et al.~\cite{7} constructed an evaluation model that evaluates a vast number of generated grasp candidates, enabling the selection of proper grasps. Mahler et al.~\cite{8} used reinforcement learning to solve the problem of clutter. Following that, they continued their work based on vacuum end effectors~\cite{9}. A grasp-first-then-recognize work-flow for grasping was proposed by Andy et al.~\cite{10}, which performs well in the stowing task. 

\textbf{Grasping algorithms utilizing point cloud data} Since PointNet~\cite{11} and PointNet++~\cite{12} have demonstrated exceptional performance in the detection, segmentation and classification tasks based on point cloud data, the types of data used in object grasping algorithms had generally transformed from RGB or RGB-D images into point cloud~\cite{13, 14, 15, 16, 17}. Simultaneously, the performance of 6D pose estimation~\cite{18, 19, 20} was also greatly improved with the invention of PointNet++, which resulted in more researchers to choose PointNet++ as their backbone of the network. 6-DOF GraspNet~\cite{21} utilized PointNet++ as its backbone to estimate the 6D poses of the target object in order to obtain its accurate position, which could increase the successful possibility of the final grasp. Similar to PointNet++, other advanced deep learning techniques were also incorporated into object grasping algorithms, such as Transformers~\cite{22}, self-Supervised learning~\cite{23, 24, 25} and transfer learning~\cite{26, 27, 28}. In addition to incorporate deep learning techniques, researchers also begun to investigate how to generate grasps on objects made of different materials. To grasp transparent objects,~\cite{29, 30} altered the structure of the gripper in order to solve the problem of inaccurate depth information for transparent objects captured by cameras. As for flexible objects,~\cite{31} constructed the prediction model to estimate the position of the objects in different time series. 

\textbf{Missing and noisy ground truth} Many typical methods from deep learning actually also play an important role in the area of the representation learning for missing labels (also called semi-supervised learning). CCGAN~\cite{32} demonstrates that the surrounding parts of the input image is able to provide the context information, which can contribute the generator to generate pixels for the missing parts. As for pseudo-label methods~\cite{33}, Entropy Minimization~\cite{34}, a strategy to prevent stop the boundary from passing through the dense data points region, provides the foundational theoretical knowledge. For example, Noisy Student~\cite{35} put forward a semi-supervised method which proposed two network models, one is called student and another is teacher, to incorporate during training.~\cite{36} is commonly regarded as one of the most seminar work in the research area of the label-noise representation learning, above which an additional constrained linear “noise” layer has been introduced. This layer adjusts the output of the network to simulate a noisy label distribution. Since then, this specific area became flourished. The experiment results from~\cite{37} strongly demonstrate that the robustness against label destruction, especially for large-scale noisy datasets, could be promoted by pre-training. Dividemix~\cite{38} came up with the idea that learning all the samples is not essential for the neural network. Network can pick up the non-confusing samples as their datasets. Furthermore, the experiments demonstrate that a staged training approach can effectively alleviate the inference caused by noisy labels~\cite{39, 40}. In addition, many scholars established high-quality datasets for limited ground truth conditions~\cite{41, 42, 43}. Despite the variety of algorithms for missing and noisy ground truth, we have not found one designed for robot grasping tasks. The propose of this paper fills a gap in this field.

\section{Problem statement}
Grasp under the certain framework can be represented as three elements, the orientation, the translation and the width of the gripper. Therefore, we define the grasp $\bm{G}$ as:
\[\bm{G}=[\bm{R}\, \bm{t}\, w],\]
where $\bm{R} \in \mathbb{R}^{3 \times 3}$ represents the orientation of the gripper, $\bm{t} \in \mathbb{R}^{3 \times 1}$ represents the center of the grasp and $w \in \mathbb{R}$ represents the appropriate grasping width of the target object. The determinant of the orientation matrix $\bm{R}$ must equal one and the inverse of it is its transpose, which is almost impossible for the network to learn. The classic solution is to decouple the orientation matrix as viewpoint classification and in-plane rotation. Then, just as shown in Figure 2, we can formulate the final grasp $\bm{G}$ as:
\[\bm{G}=[\bm{v}\, d\, r\, \bm{t}\, w],\]
where $\bm{v}$ represents the approaching vector, $d$ represents the distance between the center of the grasp and the center of the gripper and $r$ represents the in-plane rotation around the approaching axis. Besides, in order to make our grasp representation more visually understandable, we choose the most popular gripper, two-finger parallel gripper, as the example in our Figure 2. Grasp representation for other kinds of grippers can be defined in the same way.

\section{Robust loss functions}
\subsection{Loss function for missing ground truth}
Given a classification task, $C$ is defined as the number of categories in the samples. $\bm{B}$ and $\bm{\hat{B}}$ represent the neural network predictions for labeled and unlabeled samples respectively. As for samples, $\bm{A}$ represents the labels of labeled samples and $\bm{\hat{A}}$ represents the predicted labels of unlabeled samples. $N_l$ and $N_u$ are defined as the number of labeled and unlabeled samples. Similarly, $b^m_n$ and $\hat{b}^m_n$ represent the $m^\text{th}$ component of neural network predictions of the $n^\text{th}$ sample in the labeled and unlabeled samples. $a^m_n$ represents the $m^\text{th}$ component of label of the $n^\text{th}$ sample in the labeled samples and $\hat{a}^m_n$ represents the $m^\text{th}$ component of predicted label of the $n^\text{th}$ sample in the unlabeled samples. Besides, $p(m|n)=softmax(b^m_n)$ is defined as the predicted probability of $b^m_n$ and $\hat{p}(m|n)=softmax(\hat{b}^m_n)$ as the predicted probability of $\hat{b}^m_n$. $\hat{\bm{p}}(n)=[\hat{p}(1|n), \ldots, \hat{p}(m|n), \ldots, \hat{p}(C|n)]$ represents the predicted probability vector of the $n^\text{th}$ unlabeled sample.

To deal with the labeled samples, we adopt the cross entropy loss function:
\[L_w(\bm{B},\bm{A}) = -\frac{1}{N_l} \sum_{n=1}^{N_l} \sum_{m=1}^{C} a_n^m \cdot \log(p(m|n)),\]

While dealing with the situation of missing ground truth, the approach of pseudo-labels is selected. The core idea of pseudo-labels is to make use of the model itself to generate labels for unlabeled data. In particular, we evaluate the possibility of the artificial labels based on the argmax of the model’s output. A predefined threshold is set up, which can filter out interferences from low possibilities. Therefore, we can propose our preliminary loss function:
\begin{align*}
	L_p(\hat{\bm{B}}, \hat{\bm{A}}) = &-\frac{1}{N_u} \sum_{n=1}^{N_u} \mu(\max(\hat{\bm{p}}(n)) > \gamma) \\
	&\cdot \sum_{m=1}^{C} \hat{a}_n^m \cdot \log(\hat{p}(m|n)),
\end{align*}
where $\max(\hat{\bm{p}}(n))$ denotes the highest value in $\hat{\bm{p}}(n)$, $\gamma$ denotes the threshold and function $\mu(x) =
\begin{cases}
	1, & x > 0 \\
	0, & x \leq 0
\end{cases}$. While $\max(\hat{\bm{p}}(n))>\gamma$, we believe that the confidence of $\hat{\bm{p}}(n)$ is high and the corresponding loss function term will be reserved. Otherwise, the corresponding term will be discarded. However, we find that if $\hat{a}^m_n$ is a binary variable (either 0 or 1), it obtained less information in the predictions of the network. In order to enhance the generalization capabilities of the network, our new predicted probability as follows:
\begin{equation}
	\hat{s}^m_n = \xi \hat{p}(m|n) + \frac{1 - \xi}{C - 1} \left(1 - \hat{p}(m|n)\right),
\end{equation}
where $\xi$ is a constant from 0 to 1. By substituting $\hat{s}^m_n$ for $\hat{a}^m_n$, it allows the loss function term for unlabeled samples to incorporate more network prediction information, leading to increased robustness during training. Therefore, our new loss function for unlabeled samples is:
\begin{equation}
	\begin{aligned}
		L_u(\hat{\bm{B}}, \hat{\bm{A}}) = &-\frac{1}{N_u} \sum_{n=1}^{N_u} \mu(\max(\hat{\bm{p}}(n)) > \gamma) \\
		&\cdot \sum_{m=1}^{C} \hat{s}_n^m \cdot \log(\hat{p}(m|n)).
	\end{aligned}
\end{equation}
By combining the loss function of labeled and unlabeled samples, we obtain the final loss function:
\[
	L_m = \lambda_1 L_w(\bm{B}, \bm{A}) + \lambda_2 L_u(\hat{\bm{B}}, \hat{\bm{A}}),
\]
where $\lambda_1$ and $\lambda_2$ are weights.

\subsection{Loss function for missing ground truth}

\begin{figure}[!t]
	\centering
	\begin{overpic}[width=0.6\columnwidth]{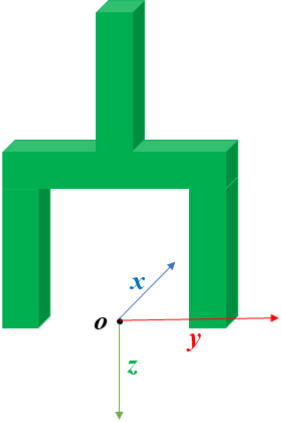}
		\put(-6.5,100){(a)} 
	\end{overpic}
	
	\vspace{0em} 
	
	\begin{overpic}[width=0.8\columnwidth]{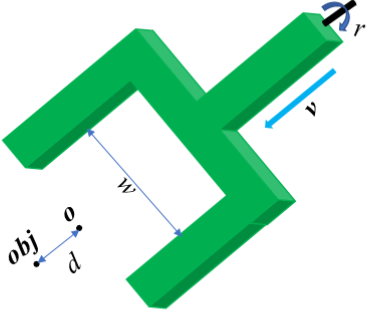}
		\put(5,70){(b)} 
	\end{overpic}
	
	\caption{The representation of the final grasp. (a) The coordinate system of the gripper. (b) Our final representation of the grasp. $\bm{obj}$ denotes the center of the object. In practical grasping scenarios, the gripper will follow the direction of $\bm{v}$ to move forward for the distance of $d$ and grasp the target object with the width $w$.}
\end{figure}

Given a classification task, $C$ is defined as the number of categories in the samples. $D$ represents samples of the dataset. Then, we define the probability of each label $c \in \{1, \ldots, C\}$ as $p(c | d) = \frac{e^{g_c}}{\sum_{j=1}^{c} e^{g_j}}$, where $g_j$ are the logits and $d \in D$. 

As for noisy ground truth, symmetric cross entropy learning algorithm~\cite{44} which is able to strike a balance between sufficient learning and robustness to noisy labels plays a pivotal role. According to this algorithm, the loss function for noisy ground truth should be defined as:
\[L_t = - \sum_{c=1}^{C} q(c | d) \log p(c | d) - \sum_{c=1}^{C} p(c | d) \log q(c | d),\]
where $q(c|d)$ should be a binary variable (either 0 or 1) and qualify $\sum_{c=1}^{C} q(c | d) = 1$. However, as mentioned in the loss function for missing ground truth, the loss function cannot bring all useful information in the noisy ground truth. Therefore, we adopt the similar idea of the construction of $\hat{s}^m_n$. Then $s(c|d)$ is defined as:
\[s(c | d) = \delta p(c | d) + \frac{1 - \delta}{C - 1} (1 - p(c | d)),\]
where $\delta$ is a constant from 0 to 1. So $s(c | d)$ is used to define our new loss function for noisy ground truth as:
\begin{align*}
	L_n = &-\alpha_1 \sum_{c=1}^{C} s(c | d) \log p(c | d) \\ 
	&- \alpha_2 \sum_{c=1}^{C} p(c | d) \log s(c | d),
\end{align*}
where $\alpha_1$ and $\alpha_2$ are weights. In order to express it clearly, we propose the following definition:
\[L_{ce} = - \sum_{c=1}^{C} s(c | d) \log p(c | d),\]
\[L_{rce} = - \sum_{c=1}^{C} p(c | d) \log s(c | d).\]
Therefore, the final loss function for the noisy ground truth can be rewritten as:
\begin{equation}
	L_n = \alpha_1 L_{ce} + \alpha_2 L_{rce} \tag{3}
\end{equation}

\section{Experiments}
In this section, we adopt the network and dataset from GraspNet-1Billion~\cite{45} to demonstrate that our loss functions can improve the performance while facing with the problem of missing or noisy ground truth. The dataset of GraspNet-1Billion contains 97,280 RGB-D images, which consists of 190 cluttered scenes with 88 different objects. Particularly, 512 RGB-D images are provided with each scene. Objects in each scene contain various grasp poses. The basic network of GrasspNet-1Billion~\cite{45} consists of three parts: ApproachNet, OperationNet, and ToleranceNet. ApproachNet is used to extract features and approaching vectors from point cloud input data, after which OperationNet utilizes the extracted features and approaching vectors to generate grasp candidates. ToleranceNet simultaneously provide the robustness and feasibility of the grasp candidates. Considering that ToleranceNet is only used for evaluating the robustness and feasibility of the grasp and is not involved in the calculation of the values of grasp candidates, we only modified the loss functions of ApproachNet and OperationNet when dealing with the conditions under limited data.

\subsection{Missing ground truth}
\begin{table*}[!t]
	\renewcommand{\arraystretch}{1.3}
	\caption{Evaluation based on ground truth of different missing ratios.}
	\label{tab:evaluation}
	\centering
	\newcolumntype{Y}{>{\centering\arraybackslash}X}
	\begin{tabularx}{\textwidth}{|Y|Y|Y|Y|Y|Y|Y|Y|Y|Y|Y|}
		\hline
		$\kappa_1$ & \begin{tabular}[c]{@{}c@{}}Me-\\thods\end{tabular} & \multicolumn{3}{c|}{Seen} & \multicolumn{3}{c|}{Unseen} & \multicolumn{3}{c|}{Novel} \\
		\cline{3-11}
		& & AP & AP$_{0.8}$ & AP$_{0.4}$& AP & AP$_{0.8}$ & AP$_{0.4}$ & AP & AP$_{0.8}$ & AP$_{0.4}$\\
		\hline
		50\% &~\cite{45} & 23.35 & 28.09 & 14.10 & 22.98 & 29.05 & 12.80 & 8.23 & 7.87 & 3.38 \\
		& Ours & \textbf{23.86} & 27.12 & \textbf{14.45} & 22.95 & \textbf{30.21} & \textbf{13.82} & \textbf{9.14} & \textbf{8.66} & \textbf{3.79} \\
		\hline
		60\% &~\cite{45} & 21.98 & 26.42 & 13.25 & 22.89 & 25.63 & 11.95 & 7.28 & 6.75 & 3.10 \\
		& Ours & \textbf{22.48} & \textbf{26.99} & 13.16 & \textbf{23.94} & \textbf{26.86} & \textbf{12.93} & \textbf{8.01} & \textbf{7.49} & \textbf{3.48} \\
		\hline
		70\% &~\cite{45} & 20.05 & 22.07 & 12.58 & 18.80 & 22.22 & 11.38 & 6.33 & 5.85 & 2.95 \\
		& Ours & \textbf{20.60} & 22.02 & \textbf{13.19} & \textbf{19.72} & \textbf{23.38} & 11.34 & \textbf{6.90} & \textbf{6.55} & \textbf{3.34} \\
		\hline
	\end{tabularx}
\end{table*}
The ground truth of one grasp is consisted of the in-plane rotation, width of the gripper, and grasping confidence score. We adopted a strategy of Missing Completely at Random (MCAR) for the ground truth, setting up constant $\kappa_1$ as the proportion of data removed in the original dataset. During training, if encountering complete ground truth, the normal loss function mentioned in the original paper~\cite{45} was utilized for training. If the ground truth is missing, the loss function proposed in section 4.1 was used.

The grasping confidence score is related to the loss function of the ApproachNet. Following the proposed loss function (as defined in Equation (2)) and the loss function from~\cite{45}, we can modify the loss function of ApproachNet in GraspNet-1Billion into the following formula:
\begin{align*}
L^A(\{c_i\}, \{s_{ij}\}) = &\frac{1}{N_{cls}} \sum_{i} L_{cls} (c_i, c_i^*) + \beta_1 \frac{1}{N_{reg}} \cdot \\
&\sum_{i} \sum_{j} c_i^* \textbf{1}(|v_{ij}, v_{ij}^*| < 5\degree) \cdot F_A,
\end{align*}

\[F_A = 
\begin{cases} 
	L_{reg} (s_{ij}, s_{ij}^*), & \text{with ground truth} \\
	L_u (s_{ij}, \hat{s}_{ij}), & \text{without ground truth}
\end{cases}\]
where $c_i$ represents the binary value for whether it is graspable or not for each point $i$, $c_{i}^*$ is regarded as 1 if point $i$ is positive and 0 if negative, $s_{ij}$ represents the predicted confidence score and $j$ means the viewpoint, $\hat{s}_{ij}$ is the predicted value based on the Equation (1), $s_{ij}^*$ is the corresponding ground truth of $s_{ij}$ and $|v_{ij}, v_{ij}^*|$ represents the angle difference between these two approaching vectors. $\beta_1$ is the constant which is usually set as 0.5. $L_{cls}$ we use here denotes a two class softmax loss and $L_{reg}$ denotes the smooth $L_1$ loss.

The in-plane rotation and the width of the gripper are associated with the loss function of the OperationNet. Following the same idea to deal with the ApproachNet, we can also modify the loss function of OperationNet in GraspNet-1Billion into the following formula: 
\begin{align*}
L^{R} (R_{ij}, S_{ij}, W_{ij}) = &\sum_{c=1}^{C} ( \frac{1}{N_{cls}} \sum_{ij} F_{O1} \\ &+ \beta_{2} \frac{1}{N_{reg}} \sum_{ij} F_{O2} \\
&+ \beta_{3} \frac{1}{N_{reg}} \sum_{ij} F_{O3}),
\end{align*}
\[F_{O1} = 
\begin{cases} 
	L^d_{cls} (R_{ij}, R_{ij}^*), & \text{with ground truth} \\
	L_u (R_{ij}, \hat{R}_{ij}), & \text{without ground truth},
\end{cases}\]
\[F_{O2} = 
\begin{cases} 
	L^d_{reg} (S_{ij}, S_{ij}^*), & \text{with ground truth} \\
	L_u (S_{ij}, \hat{S}_{ij}), & \text{without ground truth},
\end{cases}\]
\[F_{O3} = 
\begin{cases} 
	L^d_{reg} (W_{ij}, W_{ij}^*), & \text{with ground truth} \\
	L_u (W_{ij}, \hat{W}_{ij}), & \text{without ground truth},
\end{cases}\]
where $R_{ij}$ represents the rotation degree, $S_{ij}$ denotes the grasp confidence score, $W_{ij}$ are regarded as gripper width. $\hat{R}_{ij}$, $\hat{S}_{ij}$ and $\hat{W}_{ij}$ represent the predicted values of $R_{ij}$, $S_{ij}$, and $W_{ij}$ respectively. $R_{ij}^*$, $S_{ij}^*$ and $W_{ij}^*$ represent the corresponding ground truth of $R_{ij}$, $S_{ij}$, and $W_{ij}$ respectively. $L^d$ represents the mean loss for the $d^\text{th}$ binned distance. $\beta_{2}$ and $\beta_{3}$ are constants. In particular, $L_{cls}$ represents the sigmoid cross entropy loss function. 

Table I demonstrates the evaluation results utilizing various ratios of ground truth. The percentages in the first column, $\kappa_1$, represent the proportion of missing ground truth in the dataset. By comparing to its original values under complete ground truth, the average AP value of GraspNet-1Billlion~\cite{45} decrease around 20\%, which proves that the performance of the grasping algorithm is highly dependent on the ground truth. In situations where ground truth is missing, training the grasping algorithm on seen and unseen datasets does not significantly deteriorate, as the target objects in these test datasets are highly similar to the objects provided in the training dataset. However, its generalization ability is greatly affected, which is shown from the evaluation results of novel dataset. By employing our loss function, the model not only achieves an average increase of 3\% on seen and unseen datasets but also significantly enhances its generalization capability, with its performance on novel dataset increasing averagely by 15\%.

\subsection{Noisy ground truth}
\begin{table*}[t]
	\renewcommand{\arraystretch}{1.3}
	\caption{Evaluation based on ground truth of different ratios and noisy factors.}
	\label{tab:evaluation_noisy}
	\centering
	\newcolumntype{Y}{>{\centering\arraybackslash}X}
	\begin{tabularx}{\textwidth}{|Y|Y|Y|Y|Y|Y|Y|Y|Y|Y|Y|Y|}
		\hline
		$\kappa_2$  & $\epsilon$ & \begin{tabular}[c]{@{}c@{}}Me-\\thods\end{tabular} & \multicolumn{3}{c|}{Seen} & \multicolumn{3}{c|}{Unseen} & \multicolumn{3}{c|}{Novel} \\
		\cline{4-12}
		& & & \textbf{AP} & AP$_\textbf{0.8}$ & AP$_\textbf{0.4}$ & \textbf{AP} & AP$_\textbf{0.8}$ & AP$_\textbf{0.4}$ & \textbf{AP} & AP$_\textbf{0.8}$ & AP$_\textbf{0.4}$ \\
		\hline
		50\%  & 0.5 &~\cite{45} & 25.01 & 30.69 & 14.64 & 23.02 & 30.10 & 12.23 & 6.49 & 6.56 & 2.62 \\
		& & Ours & \textbf{25.28} & 30.65 & \textbf{14.94} & \textbf{23.98} & 30.08 & \textbf{12.71} & \textbf{7.43} & \textbf{7.75} & \textbf{2.95} \\
		\hline
		50\%  & 0.6 &~\cite{45} & 24.73 & 30.39 & 14.53 & 22.97 & 29.93 & 12.11 & 6.44 & 6.45 & 2.66 \\
		& & Ours & \textbf{25.05} & \textbf{30.71} & 14.46 & \textbf{23.95} & \textbf{31.79} & \textbf{12.56} & \textbf{7.47} & \textbf{7.64} & \textbf{3.00} \\
		\hline
		60\%  & 0.5 &~\cite{45} & 24.18 & 29.69 & 14.17 & 22.79 & 29.90 & 12.08 & 6.39 & 6.33 & 2.58 \\
		& & Ours & \textbf{24.58} & \textbf{30.10} & 14.16 & \textbf{23.79} & \textbf{32.00} & \textbf{12.50} & \textbf{7.52} & \textbf{7.42} & \textbf{2.92} \\
		\hline
		60\%  & 0.6 &~\cite{45} & 23.63 & 28.96 & 13.81 & 24.80 & 29.45 & 11.92 & 5.97 & 6.20 & 2.50 \\
		& & Ours & \textbf{24.14} & \textbf{29.46} & \textbf{14.19} & \textbf{26.09} & 28.58 & \textbf{12.46} & \textbf{7.14} & \textbf{7.37} & \textbf{2.84} \\
		\hline
	\end{tabularx}
\end{table*}
Following the similar idea from missing ground truth condition, the noisy ground truth was generated through changing the values of the in-plane rotation and the width of the gripper. In particular, as we did not modify the grasping confidence scores, the loss function of the ApproachNet remains unchanged. Firstly, we set up two constants, $\kappa_2$ and $\epsilon$. $\kappa_2$ represents the proportion of data modified in the original dataset. $\epsilon$ represents the extent to which the ground truth is modified. Both $\kappa_2$ and $\epsilon$ are in the range from 0 to 1. In this part of experiment, we randomly selected a propotion, $\kappa_2$, of the ground truth and multiplied them all by $\epsilon$ to achieve the effect of noisy ground truth. 

The in-plane rotation and the width of the gripper are in connection with the loss function of the OperationNet. Following the proposed loss function (as defined in Equation (3)) and the loss function from~\cite{45}, the loss function of OperationNet for the noisy ground truth in GraspNet-1Billion can be constructed as the following formula: 
\begin{align*}
L^R(R_{ij}, S_{ij}, W_{ij}) = &\sum_{c=1}^{C} ( \frac{1}{N_{cls}} \sum_{ij} (L_{ce}(R_{ij}, R^*_{ij}) \\
&+ L_{rce}(R_{ij}, R^*_{ij})) \\
&+ \eta_2 \frac{1}{N_{reg}} \sum_{ij} (L_{ce}(S_{ij}, S^*_{ij}) \\
&+ L_{rce}(S_{ij}, S^*_{ij})) \\
&+ \eta_3 \frac{1}{N_{reg}} \sum_{ij} (L_{ce}(W_{ij}, W^*_{ij}) \\
&+ L_{rce}(W_{ij}, W^*_{ij}))),
\end{align*}
the parameters setting is the same as Section 5.1. $\eta_2$ and $\eta_3$ are constants.

Table II shows the evaluation results using various ratios and noisy factors of ground truth. The first column and the second column represent the ratios of noisy ground truth $\kappa_2$ and the noisy factors $\epsilon$ respectively. By constructing a symmetric loss function, the influence of noisy ground truth was reduced on the final calculation of the loss function. This approach enhances the effectiveness of training process. The improvement in values of Table II demonstrates that the accuracy and generalization ability have both been further enhanced under conditions of noisy ground truth. On seen and unseen datasets, the performances contain an overall increase of 2\%, and on the novel datasets, the performances obtain an overall increase of 10\%.

\section{Conclusion}
In this paper, we proposed two loss functions to address the issues of missing ground truth and noisy ground truth in grasping algorithms. For missing ground truth, the pseudo-labeling approach is utilized to mitigate the problem of insufficient data where our proposed predicted category possibility method plays an essential role. For noisy ground truth, a symmetric loss function is constructed to reduce the impact of label noises on training. Through comparative experiments, we demonstrated that these two loss functions can not only enhance the accuracy of algorithm but also improve its generalization ability under the condition of limited data. For future work, we will attempt to build a dataset specifically for training grasping algorithms under limited conditions, filling a gap in this field.

\bibliographystyle{unsrt}
\bibliography{refs}

\end{document}